\documentclass{article}
\usepackage{ijcai23}

\usepackage{times}
\usepackage{soul}
\usepackage{url}
\usepackage[hidelinks]{hyperref}
\usepackage[utf8]{inputenc}
\usepackage[small]{caption}
\usepackage{graphicx}
\usepackage{amsmath}
\usepackage{amsthm}
\usepackage{booktabs}
\usepackage{algorithm}
\usepackage{algorithmic}
\usepackage[switch]{lineno}
\usepackage{CJKutf8}
\usepackage{threeparttable}    
\usepackage{makecell}
\usepackage{threeparttable}    
\usepackage{multirow}
\usepackage{diagbox}
\usepackage{amssymb}

\title{Deformation Robust Text Spotting with Geometric Prior}

\author{
Xixuan Hao$^{1,2}$
\and
Aozhong Zhang$^1$\and
Xianze Meng$^1$\And
Bin Fu$^1$
\affiliations
$^1$ ShenZhen Key Lab of Computer Vision and Pattern Recognition, Shenzhen Institute of Advanced Technology, Chinese Academy of Sciences \\
$^2$ The University of Hong Kong \\
\emails
hxxjxw@connect.hku.hk,
\{gz.zhang1, xz.meng, bin.fu\}@siat.ac.cn,
}

\begin{document}
\newcolumntype{C}[1]{>{\centering\arraybackslash}p{#1}}
\maketitle

\begin{abstract}
The goal of text spotting is to perform text detection and recognition in an end-to-end manner.
Although the diversity of luminosity and orientation in scene texts has been widely studied, the font diversity and shape variance of the same character are ignored in recent works, since most characters in natural images are rendered in standard fonts. 
To solve this problem, we present a Chinese Artistic Dataset, termed as ARText, which contains $33,000$ artistic images with rich shape deformation and font diversity. 
Based on this database, we develop a deformation robust text spotting method (DR TextSpotter) to solve the recognition problem of complex deformation of characters in different fonts. 
Specifically, we propose a geometric prior module to highlight the important features based on the unsupervised landmark detection sub-network. 
A graph convolution network is further constructed to fuse the character features and landmark features, and then performs semantic reasoning to enhance the discrimination for different characters. 
The experiments are conducted on ARText and IC19-ReCTS datasets. 
Our results demonstrate the effectiveness of our proposed method. 
\end{abstract}

\section{Introduction}
\label{section:1}
Spotting text, aiming at detecting and recognizing text in an end-to-end manner, is an emerging topic in the computer vision community, 
due to the wide application in machine translation, image retrieval and autonomous driving. 
In recent years, it has achieved significant progress due to the rapid development of deep neural networks. 
The diversity of texts is the critical challenge for text spotting tasks, and various methods \cite{xing2019convolutional,wang2020ae,liao2020mask,9525302,wang2021pan++,huang2022swintextspotter} have been proposed to handle the diversity problem of luminosity and orientation in scene texts. 
Unfortunately, we find the intrinsic diversity of characters, namely the font diversity, has been almost ignored in recent studies, where the large variance of shape of the same character in various fonts may cause significant performance degradation in reading texts. 

\begin{figure}[h!]
\centering
\includegraphics[scale=0.5]{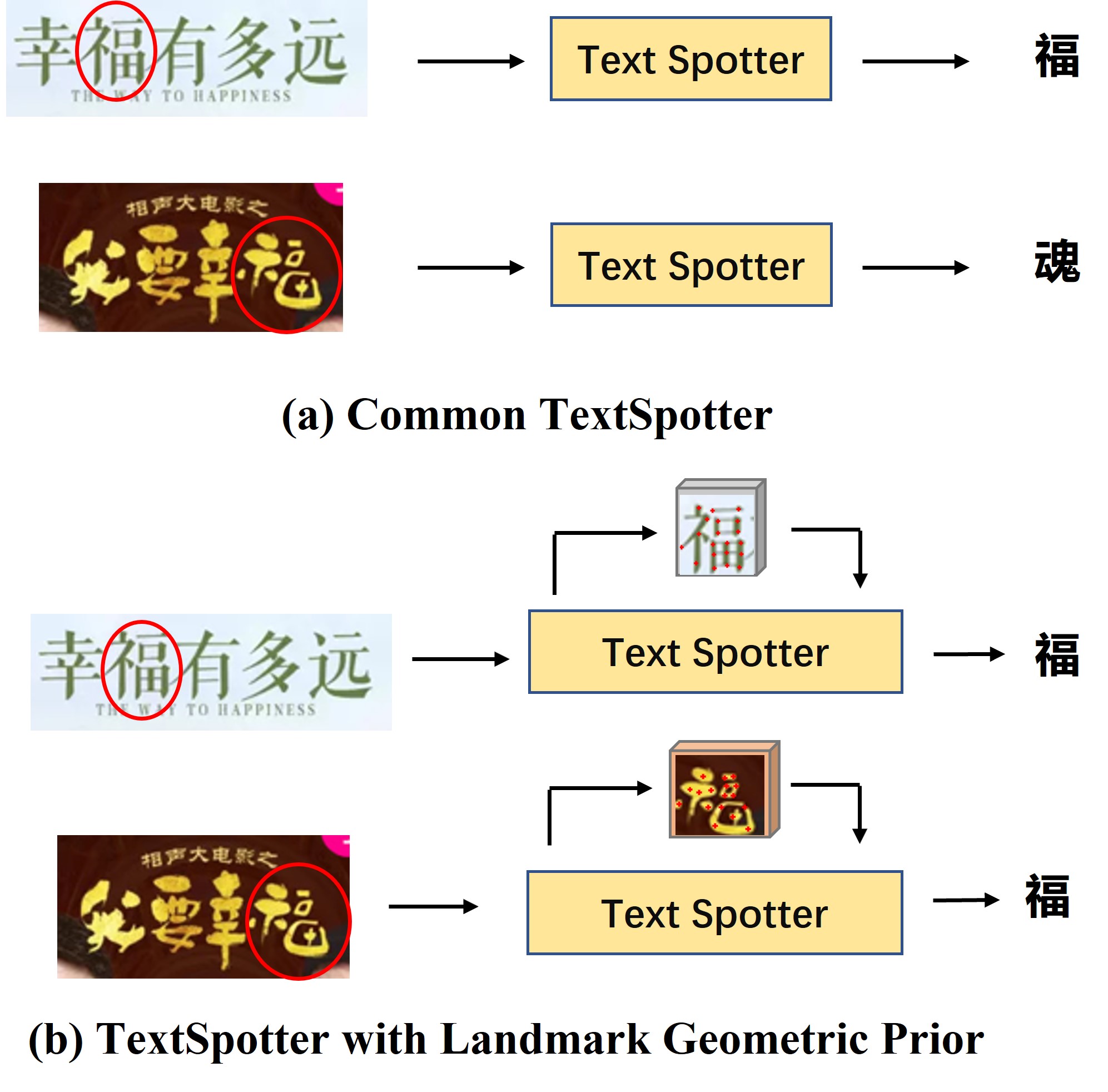}
\caption{The illustration of our motivation. (a) is the recognition result of common text spotter, (b) is the recognition result of our proposed method. For characters with rich shape deformation, (a) is tend to misrecognize the content. For example, \begin{CJK*}{UTF8}{gbsn}“福”\end{CJK*} in standard font can be recognized correctly in the top line of (a), but \begin{CJK*}{UTF8}{gbsn}“福”\end{CJK*}  in the complex font is misrecognized as \begin{CJK*}{UTF8}{gbsn}“魂”\end{CJK*} in the bottom line of (a). With geometric prior, our model can get more information about the basic architecture of characters, thus being more capable of recognizing characters with diversified fonts. For example, \begin{CJK*}{UTF8}{gbsn}“福”\end{CJK*}  in the complex font is recognized correctly in the bottom line of (b).}
\vspace{-10pt}
\label{fig:idea}
\end{figure}

\begin{table*}[h]
	\centering

	\renewcommand\arraystretch{1.5}
	\small
    	\begin{tabular}{c|c|c|c|c|c|c|c|C{1.2cm}|C{1.2cm}|c}
    		\hline
    		 & Detection & 1-NED &\multicolumn{3}{c|}{1-NED}& $N_1$ & $N_2$ & \multicolumn{3}{c}{Font Difficulty Distribution}\\
    		\hline
    		\multirow{2}{*}{A set} & \multirow{2}{*}{0.993} & \multirow{2}{*}{0.977} &simple&medium&hard& \multirow{2}{*}{8960} & \multirow{4}{*}{1923} & \multirow{2}{*}{simple} &  \multirow{2}{*}{medium} & \multirow{2}{*}{hard} \\
    		\cline{4-6}
    		&&&0.984&0.978&0.965&&&&\\
    		\cline{1-7}
    		\multirow{2}{*}{B set} & \multirow{2}{*}{0.990} & \multirow{2}{*}{0.918}  &simple&medium&hard& \multirow{2}{*}{7623} & &\multirow{2}{*}{289} & \multirow{2}{*}{299} & \multirow{2}{*}{935} \\
    		\cline{4-6}
    		&&&0.971&0.953&0.781&&&& \\
    		\hline
    	\end{tabular}
 \captionsetup{singlelinecheck=off,justification=raggedright}
	   \caption{Experiments on generalization ability for text spotting in complex font scenarios. We use the edit-distance-based metric, 1-NED, to evaluate the recognition performance. $N_1$ stands for the Number of completely recognized images. $N_2$ stands for the Number of A-right-B-wrong image pairs. Font Difficulty Distribution represents font difficulty distribution of A-right-B-wrong images.}

	\label{tab:DR} 
\vspace{-5pt}
\end{table*}

Since the font of characters in most scene text images is still close to the standard fonts, the  importance of font diversity receives limited attention.
To illustrate the above problem, we perform experiments on the generalization ability for text spotting in complex font scenarios. 
We select $100$ kinds of Chinese fonts, including simple, medium and hard categories, then split them randomly into two sets A and B, each set contains $50$ fonts (the proportion of simple, medium and hard fonts remains the same $8:11:6$ in two sets). 
We utilize A set font to generate $50,000$ text images, then split them as $40,000$ images in training set and $10,000$ images in testing set. 
Based on the testing images in A set, we randomly select fonts in B set and generate test images with the same content of testing images in A set. 
Then we optimize AE TextSpotter \cite{wang2020ae} on training images and test the recognition performance on testing images of A set and B set respectively. 
As shown in Table \ref{tab:DR}, the recognition performance has dropped by about 6\% on B testing set whose fonts are never seen before by the model, while the detection performance remains almost the same, which illustrates that the performance degradation is caused by the recognition part.
We also summarize the number of A-right-B-wrong image pairs in Table \ref{tab:DR} and analyze the image font difficulty distribution of misrecognized images, finding that
the probability of inaccurate recognition increases with font difficulty.
Therefore, the font diversity is a significant challenge for current textspotter models. 
Besides, the difficulties and challenges of font diversity are discussed in \cite{xie2022toward}, which can also support our conclusion.

In this paper, to solve the above problems, we propose a {\bf Chinese Artistic Text Dataset}, termed as ARText, which focuses on the font diversity and shape variance of characters in text spotting. 
The ARText dataset contains $33,000$ images collected on the Internet with rich shape deformation and font diversity, and has been randomly divided into training set ($30,000$ images) and testing set ($3,000$ images) for model development. 
The character-level and text-level annotations are both provided, including the 
bounding boxes and the corresponding content, which can be utilized in text detection, text recognition and text spotting tasks. 
The detailed annotation strategy and statistic information will be discussed in Section \ref{section:3}. 

Based on this dataset, we develop a deformation robust text spotting method (DR TextSpotter) to solve the recognition problem caused by complex shape deformation and font variance, as shown in Figure \ref{fig:idea}.
Specifically, motivated by the fact that the basic geometric relations of the same character are preserved in different fonts, we introduce the geometric prior to highlight the important features through the unsupervised landmark detection sub-network. 
Moreover, due to the sparse nature of landmark-based features, we embed all features into graph space, then the graph convolution network is utilized to perform semantic reasoning. 
With the geometric prior, the graph convolution can effectively encode landmark information, which  will enhance the discrimination of inter-character features and the stability of intra-character features, thus improving the recognition ability for characters with diversified fonts. 
To demonstrate the effectiveness of our proposed method, we conduct experiments on ARText dataset and widely-used IC19-ReCTS dataset.  
Experimental results show our method achieves state-of-the-art performance on IC19-ReCTS dataset \cite{zhang2019icdar}, and outperforms its counterparts about 1.5\% in terms of the edit-distance-based evaluation metric (1-NED) on our ARText dataset.

The main contributions are summarized as follows: 1) We construct a Chinese artistic text dataset, namely ARText, which is a Chinese dataset focusing on the font diversity and shape variance of characters in the text spotting task. 2) We propose a novel network architecture Deformation Robust TextSpotter, which employ geometric prior and graph neural network to enhance the discrimination of inter-character features and the stability of intra-character features.
3) Experimental results demonstrate the effectiveness of DR TextSpotter. 
On ARText dataset, DR TextSpotter achieves state-of-the-art performance, surpassing other methods by about 1.5\% in terms of  1-NED.

\begin{table*}[h]
	\centering

	\renewcommand\arraystretch{1.5}
	\small
	\begin{tabular}{l|c|c|c|c|c|c|c|r}
		\hline
		 \diagbox{Difficulty}{Types} &TV Series & Movie & Variety Show & Comic & Children & Documentary & UGC & Total\\
		\hline
		Simple &    68&  6    &  16  &  19  &  0    &  0   &  7955 & 8064\\
		\hline
		Medium &  4527&  2158 &  871 & 657  &  775  &  907 &  5212 & 15107\\
		\hline
		Hard &  3499&   2450 &  1097 & 1431 &  727  & 539  &  86  & 9829\\
		\hline
		 Total&      8094&   4614 &  1984 & 2107 &  1502 & 1446 & 13253 & 33000\\
		\hline
	\end{tabular}
 	\caption{Statistics for the ARText dataset. 
 }
	\label{tab:dataset-statistics}  
	\vspace{-5pt}
\end{table*}

\section{Related Work}
\label{section:2}
\subsection{Scene Text Spotting}
\label{section:2.1}

Scene Text Spotting\cite{li2017towards,liu2018fots,Lyu_2018_ECCV,qin2019towards,xing2019convolutional,liu2020abcnet,wang2020ae,feng2019textdragon,he2018end,baek2020character,wang2021pan++,huang2022swintextspotter} is commonly composed of two subtasks: text detection \cite{xu2022morphtext,su2022textdct,xu2022arbitrary} and text recognition \cite{li2022dual,li2021character,peng2022recognition}. 
In the past years, the rise of deep learning and the great advances in object detection have promoted the prosperity of this field.
\cite{li2017towards} first used RoI Pooling to correlate detection and recognition features via end-to-end training, but it can only handle horizontal and centralized text. 
Inspired by the Mask R-CNN, Mask TextSpotter\cite{liao2020mask} addressed text of various shapes with the help of instance segmentation. 
ABCNet\cite{9525302} introduced Bezier curves and new sampling methods (BezierAlign) to  localize oriented and curved scene text.AE TextSpotter \cite{wang2020ae} added  linguistic representation to text detection to solve the ambiguity problem and designed a novel text recognition module to achieve fast recognition. 
PAN++ \cite{wang2021pan++} utilized kernel representation to achieve arbitrarily-shaped text spotting.
Recently, as the Transformer architecture \cite{vaswani2017attention} begins to sweep into the Computer Vision community from Natural Language Processing area, there are also some transformer-based text spotting works emerging. For example, Swin TextSpotter \cite{huang2022swintextspotter} first introduced Swin Transformer \cite{liu2021swin} backbone into end-to-end text spotting and achieves promising results. \cite{xie2022toward} used cross-attention to merge the encoded image feature and corner map feature.

In these methods, however, the importance of geometric prior in text spotting was ignored. In this work, we introduce landmarks as geometric information to achieve deformation robust text spotting.

\subsection{Geometric Prior}
\label{section:2.2}
A vast amount of literature indicates that geometric priors can help deep learning models improve performance, such as shape completion\cite{dai2017shape}, position embedding\cite{vaswani2017attention} and landmark detection.
Landmark detection is often used as a kind of structural information to assist network learning. For instance, there are many traditional methods on human faces such as active appearance models \cite{cristinacce2008automatic}, template-based methods \cite{pedersoli2014using} and regression-based methods \cite{cao2014face}. 
After the rise of deep learning, more accurate landmark localization and better performance can be achieved \cite{wu2017facial,xia2022sparse,li2022towards}.
In addition to human faces, landmark detection is also widely applied to Camera Localization \cite{do2022learning}, medical image analysis \cite{yin2022one}, 3D Face Reconstruction \cite{wood20223d}.
In this work, We regard landmarks as geometric information of characters to improve model generalization.

\subsection{Graph Reasoning}
\label{section:2.3}
Graph-based approaches are efficient relational reasoning methods, which have become popular in various deep learning tasks in recent years. \cite{2016Semi} first proposed graph convolution neural networks and applied them to semi-supervised learning. 
Owing to the ability to capture global information through graph propagation, graph reasoning is introduced for visual understanding tasks \cite{yao2021non,zhao2021graph,2020Graph,chen2019graph}.
Moreover, \cite{li2018beyond} and \cite{chen2019graph} adopted graph convolution reasoning to build an end-to-end module for reasoning between disjoint and distant regions. 
Inspired by these works, we employ the reasoning power of graph convolutions to fuse character features with corresponding landmark features.

\section{Chinese Artistic Dataset}
\label{section:3}
Our proposed ARText dataset consists of $33,000$ Chinese artistic images collected from the Internet. The data source, annotation details, data statistics are described below.

\textbf{Data Source:} 
We collect our ARText images all from the Internet, trying to cover a wide content range. 
Then we impose data cleaning manually on them, removing images with low-resolution or inappropriate content. Finally, we categorize these images into different types as UGC (User Generated Content),  Movies, TV Series, Children, Comic, Variety Show and Documentary.

\textbf{Annotation Details:} In each annotation, we provide text-level bounding box and character-level bounding box with their contents. The format of the bounding box includes four corner points of the orientated bounding box. In one image, there may be multiple texts in different font sizes, but we only focus on the main part and ignore some small texts. 



\begin{figure*}[h!]
\centering
\includegraphics[scale=0.5]{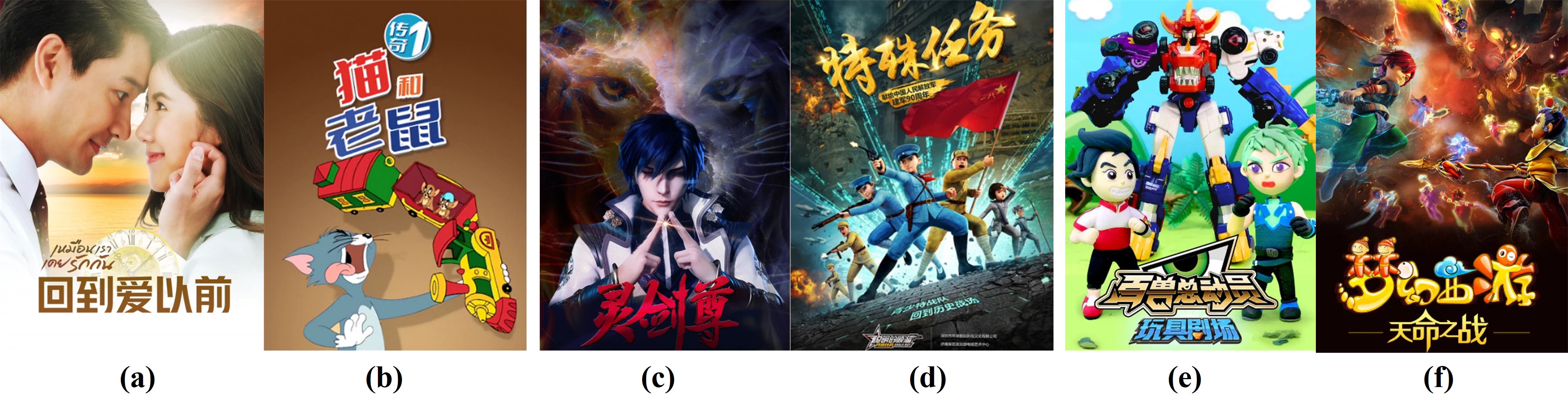}
\caption{Examples of image difficulty categories. (a)(b) are simple, (c)(d) are medium, (e)(f) are hard.}
\label{fig_difficulty_category}
\end{figure*}

\begin{figure*}[h!]
\centering
\includegraphics[scale=0.6]{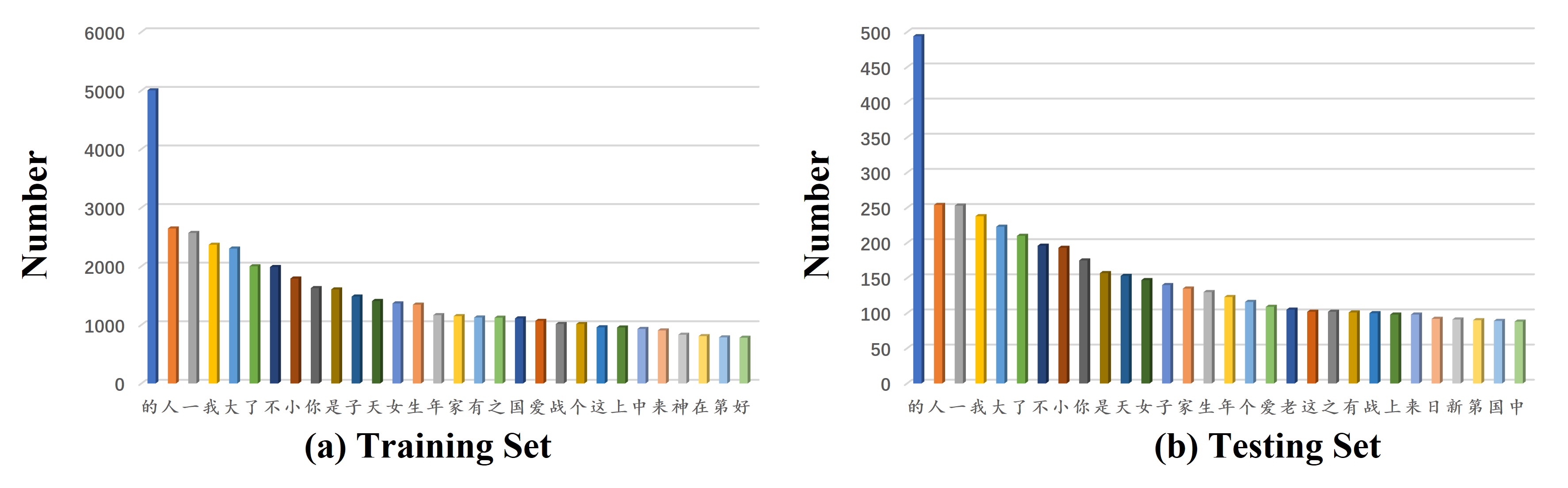}
\caption{Statistics of top-30 frequent characters in the training set (a) and testing set (b).}
\label{fig:character-frequency}
\end{figure*}

\begin{figure}[h!]
\centering
\includegraphics[scale=0.5]{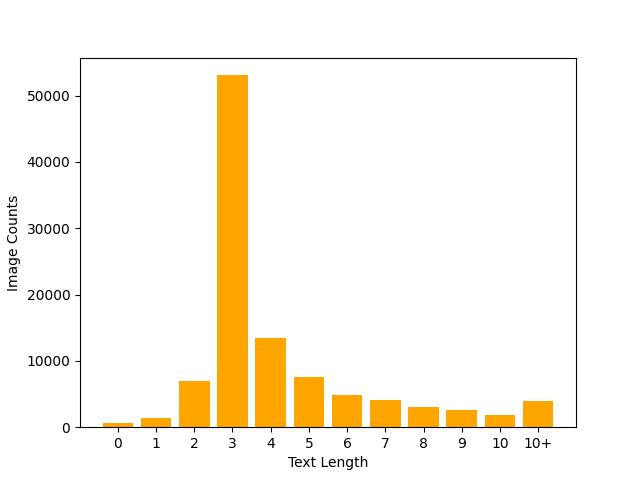}
\caption{Text length distribution of ARText.}
\label{fig:text-length}
\end{figure}

\textbf{Dataset Analysis:} The whole dataset includes $33,000$ images with 3563 different kinds of Chinese characters. The dataset statistics can be seen on Table \ref{tab:dataset-statistics}. We label dataset images into three categories (Simple, Medium, Hard) manually, according to the recognition difficulty standard, as described in Table \ref{tab:font_difficulty_standard}.
The examples are shown in Figure \ref{fig_difficulty_category}.  
Moreover, we calculate the occurrence frequency of each character in both training set and testing set. The top-$30$ characters are presented in Figure \ref{fig:character-frequency}, and the character frequency distribution of training and testing set basically remains the same. We also calculate the text length distribution of the whole dataset, which is shown in Figure \ref{fig:text-length}. Note that there might be multiple text instances in one image, so the image counts can be larger than ARText image number.

\begin{table}[h]
	\centering

	\small
    	\begin{tabular}{m{1cm}<{\centering} | m{6cm}<{\centering} }
    		\hline
    		Font & Standard\\
    		\hline
    		Simple & The font has no decoration, and its glyph has almost no deformation compared to standard fonts. \\
            \hline
    		Medium & The font has no decoration, and its glyph has deformation compared to standard fonts.\\
    		\hline
    		Hard & The font has rich decoration, and its glyph has deformation compared to standard fonts.\\
    		\hline
    	\end{tabular}
     	\caption{Font classification standard. }
      	\label{tab:font_difficulty_standard}  
     \vspace{-10pt}
\end{table}

\section{Methodology}
\label{section:4}
\subsection{Overall Pipeline}
\label{section:4.1}

\begin{figure*}[h!]
\centering
\includegraphics[scale=0.45]{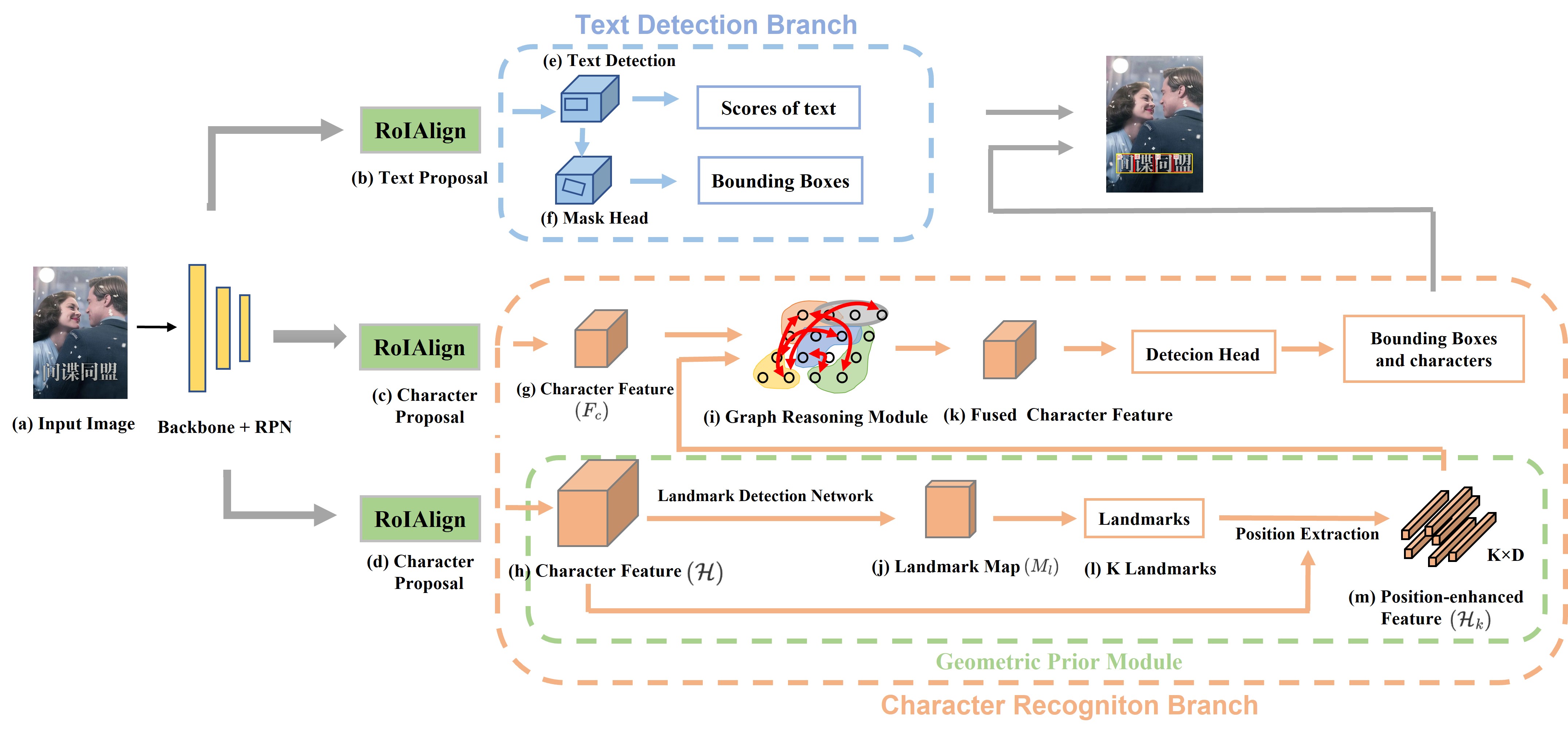}
\captionsetup{singlelinecheck=off,justification=raggedright}
\caption{Overall Architecture of DR TextSpotter, which mainly consists of Feature Extraction stage (Backbone + RPN), Proposal Generation and Extraction stage (ROIAlign), Text Detection Branch, and Character Recognition Branch. 
}
\label{fig:overall-architecture}
\vspace{-5pt}
\end{figure*}

The overall pipeline of our Deformation Robust TextSpotter is shown in Figure \ref{fig:overall-architecture}, which mainly consists of Feature Extraction stage, Proposal Generation and Extraction stage, Text Detection Branch, and Character Recognition Branch. 
We implement the Region Proposal Network (RPN) \cite{ren2015faster} on top of ResNet-50 \cite{he2016deep} as the Feature Extraction stage to predict text line proposals and character proposals. 
The ROIAlign is utilized as Proposal Generation and Extraction stage to crop the resulting proposal regions and reshape them into the predefined size. 
For the text proposal, Text Detection Branch predicts the probability of this region and further refines the text location with the oriented bounding boxes. 
For the character proposal, RoIAlign\cite{he2017mask} transformed the character proposal to character feature maps $F_c$ and $\mathcal{H}$ with different sizes, then pass them into the Character Recognition Branch. 
The Geometric Prior Module (GPM) is proposed to generate a predefined number of landmark features $\mathcal{H}_k$ according to the predicted positions. 
The landmark features and the character features $F_c$ are passed to the Graph Reasoning Module (GRM), which performs graph convolution to reason semantic relations under geometric prior. 
The resulting features will be further fed to Detection Head to predict characters and bounding boxes. 
Finally, we assemble characters according to the Intersection-over-Union (IOU) between the predicted text line (generated by the detection branch) and the character bounding box. 

\subsection{Geometric Prior Module}
\label{section:4.2}
As we discussed in the previous section, although the large shape variance and font diversity exist in the same kind of characters, the basic token relations will remain unchanged. 
Based on this motivation, we propose the Geometric Prior Module (GPM) to highlight the critical shape information in character features.

In this work, the central insight of GPM is that the geometric relations in each character can be summarized by a set of key points. 
Based on this insight, we build a landmark detection sub-network on top of the character feature patches $\mathcal{H} \in R^{70\times 70 \times D}$ (D denotes feature dimension), which consists of three $3\times 3$ convolutional layers (the first one layer is with stride 2 and the last two layer is with stride 1). 
The first two convolutional layers are followed by Batch Normalization and ReLU, while the last convolutional operator followed by the softmax function predicts the landmark map $M_l \in R^{35\times 35\times K}$ (K is the number of landmarks), where the position of the maximum value in each channel $m$ indicates the position of the $m$-th landmark.  Then we use soft-argmax operation to get the corresponding landmarks, and extract features at landmark position from the original character feature $\mathcal{H} \in R^{70\times 70 \times D}$, obtaining the landmark features $\mathcal{H}_k\in R^{K\times D}$. 

\begin{figure}[h!]
\centering
\captionsetup{singlelinecheck=off,justification=raggedright}
\includegraphics[scale=0.4]{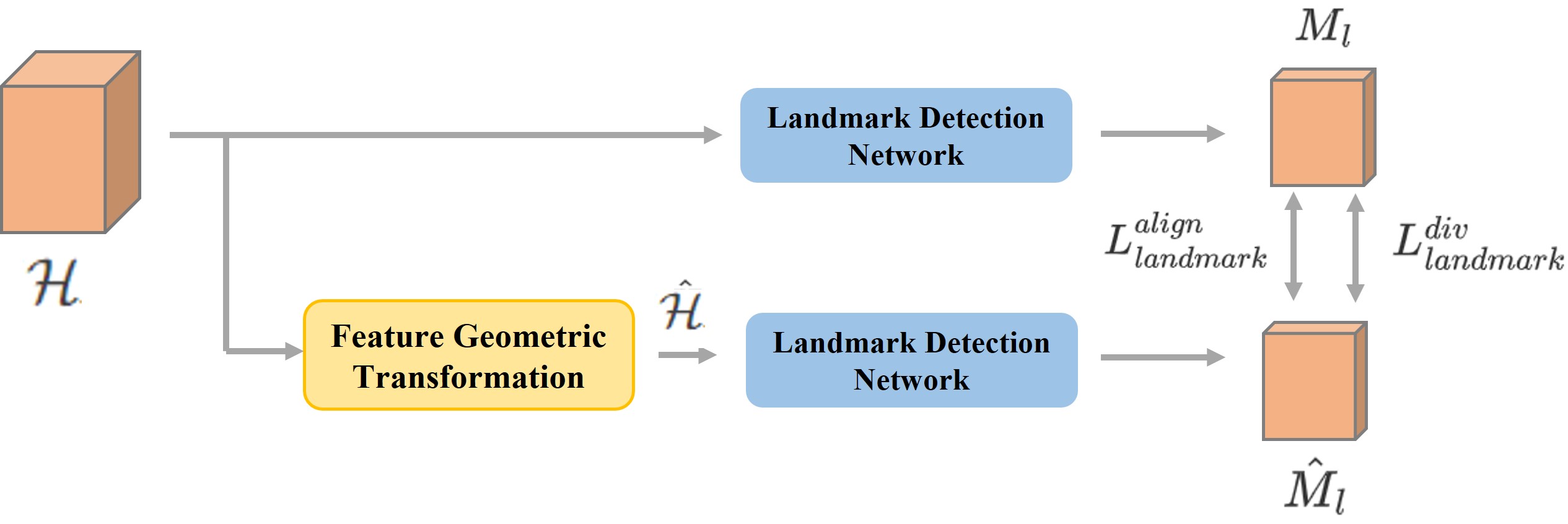}
\caption{Geometric Prior Module in Unsupervised Training Process. }
\label{Geometric Prior Module}
\end{figure}

Due to the lack of landmark annotation in the existing text spotting dataset, the training process of GPM is a challenging task. As shown in Figure \ref{Geometric Prior Module}, 
We develop an unsupervised training strategy to remedy this problem. 
Since the landmarks are the most remarkable positions for each character, we may expect that the GPM will find the same landmarks after random perturbations. 
In other words, the landmark has the characteristic of transformation equivariance. 
Therefore, once we obtain the character feature patch $\mathcal{H} \in R^{70\times 70 \times D}$, we generate another feature patch $\hat{\mathcal{H}} \in R^{70\times 70 \times D}$ by three kinds of geometric transformation $g$: rotation, translation and scaling. 
The feature maps $\mathcal{H}$ and $\hat{\mathcal{H}}$ are both passed to the Landmark Detection Network to generate landmark maps $M_l$ and $\hat{M}_l$, respectively.

We apply unsupervised optimization objections to train GPM. The first optimization target is the alignment loss:

\begin{equation}
\label{alignloss}
    L_{landmark}^{align} = \frac{1}{K} \sum_{r=1}^{K}{\sum_{uv} {||u-g(v)||^2 M_l(u|\mathcal{H}, r) \hat{M_l}(v|\hat{\mathcal{H}}, r)} }
\end{equation}
where $u$ and $v \in  \{1\cdots H\}\times \{1\cdots W\} $ indicates location, $g$ denotes feature geometric transformation. $K$ denotes the number of landmarks. $M_l(u|\mathcal{H},r)$ is the landmark maps. $r$ denotes the channel of landmark map $M_l$.

In practice, implementing Equation \ref{alignloss} is computation inefficient, so we decompose the loss as a format of linear time complexity \cite{thewlis2017unsupervised}:

\begin{equation}
\label{alignloss_linear}
\begin{split}
\sum_{u} {||u||^2 M_l(u|\mathcal{H}, r)} + \sum_{v}{||g(v)||^2 M_l(v|\hat{\mathcal{H}},r)}\\ -2 \ (\sum_{u}{u M_l(u|\mathcal{H},r)})^T \cdot (\sum_{v}{g(v) M_l(v|\hat{\mathcal{H}},r)})
\end{split}
\end{equation}

With the Equation \ref{alignloss_linear}, the landmark detection network may come to a trivial solution, where all landmarks converge into a single position. 
To solve this issue, we apply the second optimization objection, the diversity loss \cite{thewlis2017unsupervised}:
\begin{equation}
    L_{landmark}^{div} = K-\sum_u{\max_{r=1,...,K} {M_l(u|\mathcal{H}, r)}} 
    \label{divloss}
\end{equation}

This loss imposes penalties to the overlap of the max value of landmark probability map, thus preventing the network from learning K identical landmarks.
In practice, we utilize a downsampling operation (4×4 average pooling) before applying the diversity loss to encourage further landmark extraction.

\begin{figure*}[h!]
\centering
\includegraphics[scale=0.4]{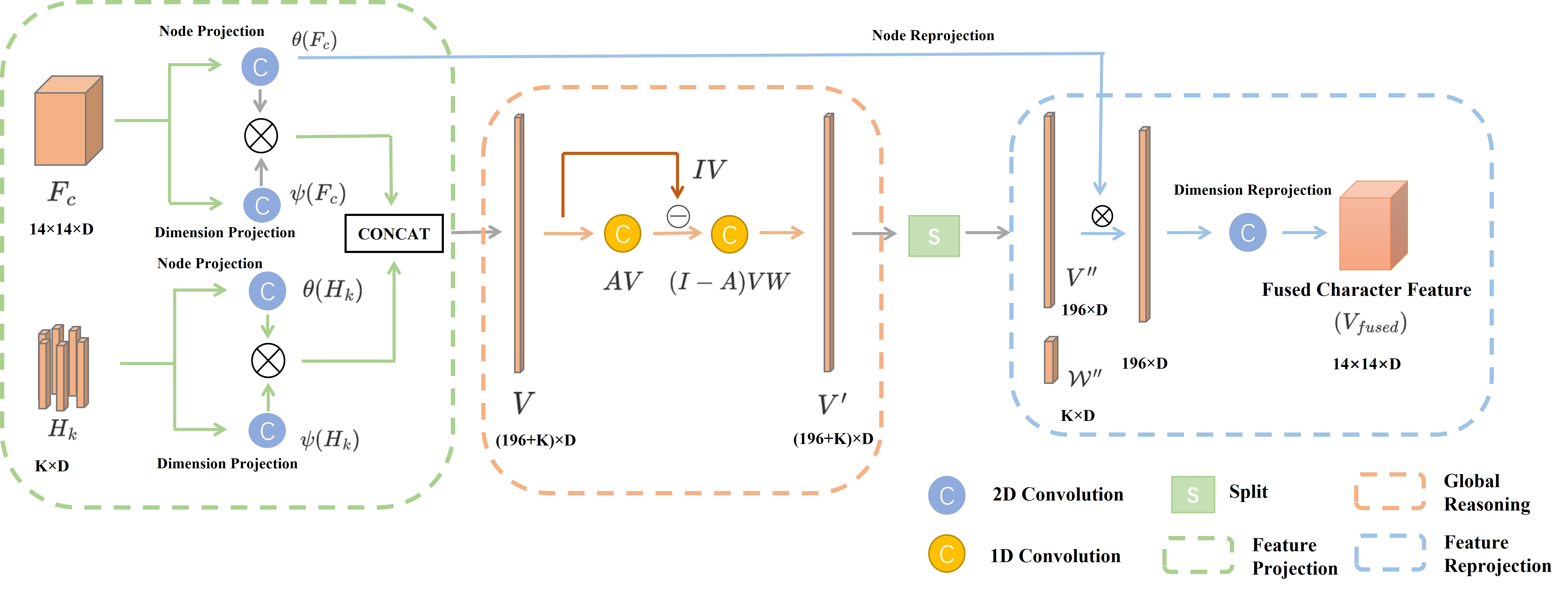}
\captionsetup{singlelinecheck=off,justification=raggedright}
\caption{The architecture of GRM. GRM includes three steps: Feature Projection, Global Reasoning and Feature Reprojection. For simplicity, the convolution layers are utilized to implement Projection, Global Reasoning and Reprojection operations.}
\label{fig:gffm}
\end{figure*}

\subsection{Graph Reasoning Module}
\label{section:4.3}
Since the landmark feature $\mathcal{H}_k \in R^{K\times D}$ is the non-grid feature, we construct the graph structure to perform global semantic reasoning in the interaction space. 
Inspired by GloRe unit \cite{chen2019graph}, the process of our GRM can be summarized in three steps, as shown in Figure \ref{fig:gffm}.

{\bf 1) Feature Projection:}
Firstly, we need to transform character features $F_c$ and landmark features $\mathcal{H}_k$ to interaction space. 

For input features X $\in \mathbb{R}^{L\times C}$, our goal is to find a map function to project X to the new features $Z = f(X) \in \mathbb{R}^{N \times C'} $ in the interaction space, where $C$ and $C'$ are the feature dimension, and we set $C=C'$ in this module. $L=W×H$ is the position dimension and $N$ is the number of features in interaction space.

\begin{equation}
z_i^{q} = w_i^{q} X = \sum_{\forall j}{w_{ij}^qx_j} 
\end{equation}
with learnable weights $W_p^q = [w_1^q,...,w_N^q], w_i^q \in \mathbb{R}^{L}; X = [x_1,...,x_N], x_j \in \mathbb{R}^{C}$, where $q \in \{c, k\}$ denotes the weight for $F_c$ or $\mathcal{H}_k$. 
In practice, we implement $f(X)$ into two functions, $\psi(X;W^q_\psi)$ for dimension projection and  
$W_p^{q}=\theta(X;W_\theta)$ for node projection. 
\begin{equation}
 f(X) =   \theta(X;W_\theta^q) \times \psi(X;W_\psi^q)
\end{equation}
The function $\psi(\cdot)$ and $\theta(\cdot)$ are implemented by a $1\times 1$ 2D convolution layer respectively. “$\times$” denotes the matrix multiplication operation. 

After projecting $F_c \in \mathbb{R}^{14\times 14\times D}$ and $H_k\in \mathbb{R}^{K\times D}$ to the interaction space, we concatenate them together and get $V\in \mathbb{R}^{(196+K)\times D}$.

2) {\bf Global Reasoning:} 
We perform semantic reasoning to embed geometric prior into features through graph convolution \cite{chen2019graph}
\begin{equation} 
U=GZW_r=(I-A)ZW_r \label{graph-equation}
\end{equation}
where $G$ and $A$ stand for adjacent matrix, $W_r$ is the weight matrix for state updating, which can be achieved by two 1D convolutions aligning both channel and node direction \cite{chen2019graph} as shown in Figure \ref{fig:gffm}. 

After global reasoning, we get $V'\in \mathbb{R}^{(196+K)\times D}$, then we split the feature $V'$ into the character feature $V''$ and landmark feature $\mathcal{W}''$. 
Finally, we reproject character feature $V''\in \mathbb{R}^{196\times D}$ back to coordinate space.

3) {\bf Feature Reprojection:} 
To make the output features of GRM can be well utilized in the subsequent recognition, the last step is to project the output features back to the original space. 
Reprojection can be formulated as 
\begin{equation}
y_i = \phi_i V'' = \sum_{\forall j}{\phi_{ij}v''_j} \label{XX}
\end{equation}
where $\Phi = [\phi_1,...,\phi_L], \phi_i \in \mathbb{R}^{N}$. 
In practice, we implement the above reprojection function into two functions, $\sigma$ for dimension reprojection and  $\delta$ for node reprojection: 
\begin{equation}
 h(V'') =  \sigma(\delta \times V'';W_\sigma) 
\end{equation}
The function $\sigma(\cdot; W_\sigma)$ is implemented by a $1\times 1$ 2D convolution layer and $W_\sigma$ denotes the learnable weight. We set $\delta=(W_p^c)^T$ (i.e. reuse the projection weight generated in the first step) to save computation.

\subsection{Optimization}
\label{section:4.4}
The training process of DR TextSpotter is divided into three stages: 
In the first stage, we optimize the backbone network, Region Proposal Network (RPN), Text Detection Branch (TDB) and Character Recognition Branch (CRB) by a joint loss of $L_{TDB}$ and $L_{CRB}$. 
In the second stage, we train the Geometric Prior Module (GPM) by $L_{landmark}^{align}$ and $L_{landmark}^{div}$. 
In the third stage, we fix the weight of GPM and remove the weight of the classification layer in the Detection Head of Character Recognition Branch (CRB), then finetune the whole network.

Correspondingly, the loss function can be divided into two parts: 1) The detection and recognition loss function. 2) The landmark detection loss function. 

\textbf{The detection and recognition loss function:}  $L_{Spotting}$ is a multi-task loss function which is used to optimize the backbone network, RPN, TDB and CRB (without GPM).

\begin{equation}
L_{Spotting} = L_{RPN}+L_{TDB}+L_{CRB}
\end{equation}
where $L_{RPN}$ is loss function of RPN as defined in \cite{ren2015faster}. $L_{TDB}$ and $L_{CRB}$ are loss functions for TDB and CRB (without GPM) respectively, which can be formulated as
\begin{equation}
L_{TDB} = L_{TDB}^{cls}+L_{TDB}^{bbox}+L_{TDB}^{mask}
\end{equation}
\begin{equation}
L_{CRB} = L_{CRB}^{cls} + L_{CRB}^{bbox} 
\end{equation}
where $L_{TDB}^{cls}$ and $L_{CRB}^{cls}$ are cross-entropy loss for text/non-text classification and character classification, respectively. 
$L_{TDB}^{bbox}$ and $L_{CRB}^{bbox}$ are smooth $L1$ loss for bounding box regression in TDB and CRB respectively. $L_{TDB}^{mask}$ is the binary cross-entropy loss of mask head in TDB.
\par
\textbf{The landmark detection loss function}: $L_{landmark}$ is used to implement equivariance constraint for landmark detection network
\begin{equation}
L_{landmark} = L_{landmark}^{align} + \lambda L_{landmark}^{div} 
\end{equation}
where $L_{landmark}^{align}$ is the probability map loss to learn equivariance constraint, while $L_{landmark}^{div}$ is the diversity loss to prevent the network from learning $K$ identical copies of the same landmark, as defined in Section \ref{section:4.2}.
We empirically set $\lambda=50$ 
to balance the above two loss functions.

\section{Experiments}
\label{section:5}
In this section, we first introduce the datasets and metrics, then compare our model with existing state-of-the-art methods. Finally, we conduct ablation studies to verify the effectiveness of different modules.

\subsection{Datasets}
 
\textbf{Chinese Artistic Text dataset}: To solve the problems about the font diversity and shape variance in text spotter, we propose the Chinese Artistic Text dataset, which provides $30,000$ text images for training and $3,000$ images for testing. The detailed description is provided in Section \ref{section:3}.

\textbf{IC19-ReCTS dataset}: IC19-ReCTS dataset \cite{zhang2019icdar} is proposed in ICDAR 2019 Robust Reading Challenge on Reading Chinese Text on Signboard. 
There are $25,000$ labeled images, including $20,000$ images for training and $5,000$ images for testing.

\subsection{Implementation Details}
The parameter settings in the backbone, RPN, and Text Detection Branch are the same as AE TextSpotter \cite{wang2020ae}. 
The magnitudes of geometric transformations in GPM are randomly selected by: Rotation $\in (0, 2\pi)$, Translation $\in (-0.2, 0.2)$, and Scaling $\in (0.7, 1.3)$. In GRM, the number of landmarks K is set to 16, D is set to 256, the number of nodes and dimensions in interaction space is set the same as the original coordinate space.
In the training process, we first optimize the backbone network, RPN, TDB, and CRB with batch size $8$ on $4$ Quadro RTX-8000 GPUs for $12$ epochs. 
Secondly, we optimize the Geometric Prior Network (GPM) and freeze other layers. 
Finally, we remove the weight of classification layers in the Detection Head of CRB, then finetune the whole network. 
We optimize all models using stochastic gradient descent (SGD) with a weight decay of $1 \times 10^{-4}$ with the momentum $0.9$. 
The initial learning rate is $0.02$, which will be divided by $10$ at epoch $8$ and $10$.  
The random scale, random horizontal flip, and random crop are utilized as data augmentation strategies in the training stage.

\subsection{Metrics}
The Precision, Recall and F-measurement are computed based on detection results.
Moreover, we use normalized edit distance (NED) based metric to evaluate end-to-end recognition performance, which is defined as: 
\begin{equation}
\label{1-ned}
1\mbox{-}NED = 1-\frac{1}{N}\sum_{i=1}^{N}{\frac{L(T_i,\hat{T_i})}{max(|T_i|,|\hat{T_i}|)}}
\end{equation}
where $L$ denoted the Levenshtein Distance, $T_i$ represents the predicted text and $\hat{T_i}$ represents the corresponding ground truth, $|T_i|$ and $|\hat{T_i}|$ represent the length of corresponding texts, $N$ is the total number of text lines.

\subsection{Comparisons with State-of-the-Art Methods}
To further demonstrate the effectiveness of the proposed DR TextSpotter, we compare it with existing state-of-the-art methods on our ARText dataset and widely-used IC19-ReCTS dataset \cite{zhang2019icdar}, respectively. 
We choose AE TextSpotter \cite{wang2020ae}, Mask TextSpotter v3 \cite{liao2020mask}, ABCNet v2 \cite{9525302} and SwinTextSpotter \cite{huang2022swintextspotter} as the representative methods. 
For the ARText dataset, we utilize the officially released codebase to train and evaluate each model. 
As shown in Table \ref{tab:result-poster}, our DR Textspotter outperforms baseline (AE TextSpotter without LM) with about 2.5\% and outperforms existing state-of-the-art method (SwinTextSpotter) about $1.5\%$ in terms of normalized edit distance. 
It is also worth mentioning that, with the language model, the performance of AE TextSpotter is slightly dropped, since the length of texts in ARText dataset is much shorter than the traditional scene text dataset. 
As shown in Table \ref{tab:result-ic19}, we validate our DR TextSpotter on public-available IC19-ReCTS \cite{zhang2019icdar} dataset, where our proposed model achieves the best record $73.20\%$ in terms of 1-NED.
We also conduct experiments on English dataset, the results are shown in Section \ref{section:res-on-engdata}.

\begin{table}[bt]
		\centering
		\label{tab:camvid}
	    \renewcommand\arraystretch{1.5}
	    \small

        \setlength{\tabcolsep}{1.3mm}
	\begin{threeparttable}
	\begin{tabular}{c|c|c|c|c|c|c}
        \hline
		\multirow{2}{*}{Methods} & \multicolumn{3}{c|}{Detection} & E2E  & \multirow{2}{*}{FPS} & \multirow{2}{*}{Params}  \\
		\cline{2-5}
		 & P & R & F & 1-NED &&  \\
		\hline
		Mask TS v3 &85.65 & 62.55 & 72.30 & 54.82&2.5&57.43M \\
		\hline
		ABCNet v2 & 88.62 & 87.37 & 87.99 &66.72 &10.0&48.16M \\
		\hline
		AE TS w/o LM & 88.15 & 89.13 & 88.64 & 72.12&1.8&256.0M \\
		\hline
		AE TS & 87.97&	88.97&	88.47&	72.06 & 2.5 & 224.7M\\
    		\hline
		SwinTS* & 92.44 & \textbf{89.82} & 90.15 &	73.22 & 0.2 & 225.4M\\
		\hline
		DR TS (Ours) & \textbf{92.48}&	88.67 &	\textbf{90.53} &	\textbf{74.69}  & 1.5 & 236.4M\\
		\hline
	\end{tabular}
      \end{threeparttable}
      \caption{Results on ARText dataset. "TS" denotes TextSpotter, while "LM" denotes the language module, and E2E means end-to-end text spotting result. * denotes that, to ensure a fair comparison, we make the parameters of Swin TextSpotter roughly equivalent to our model, by modifying pooler resolution and adding some convolution layers.}
      \label{tab:result-poster}  
      \vspace{-10pt}
\end{table}

\begin{table}[bt]
		\centering
		\label{tab:cocostuff}
	    \renewcommand\arraystretch{1.5}
	    \small

        \setlength{\tabcolsep}{1.3mm}
	\begin{threeparttable}
	\begin{tabular}{c|c|c|c|c|c|c}
        \hline
	\multirow{2}{*}{Methods} &\multicolumn{3}{c|}{Detection} & E2E  & \multirow{2}{*}{FPS} & \multirow{2}{*}{Params}  \\
		\cline{2-5}
		 & P & R & F & 1-NED&&    \\
		\hline
		Mask TS v2 &89.30 & 88.77 & 89.04 & 67.79 &2.0&80.13M\\
		\hline
		ABCNet v2 &  93.60 & 87.50 & 90.40 & 62.71 &10.0&48.16M\\
		\hline
		AE TS w/o LM & 91.54&	90.78 &	91.16&	71.10 & 2.5 & 224.7M\\
		\hline
		AE TS &92.60&	\textbf{91.01} &	91.80&	71.83 &1.8&256.0M\\
  		\hline
		SwinTS* & 94.15 &	87.19 &	90.57 &	72.58 &0.2&225.4M\\
		\hline
		DR TS (Ours) & \textbf{94.47}&	90.06&	\textbf{92.21}&	\textbf{73.20} & 1.5 & 236.4M\\
		\hline
	\end{tabular}
      \end{threeparttable}
      \caption{Results on IC19-ReCTS dataset. “TS” denotes TextSpotter, while “LM” denotes language module, and E2E means end-to-end text spotting result. * denotes that, to ensure a fair comparison, we make the parameters of Swin TextSpotter roughly equivalent to our model, by modifying pooler resolution and adding some convolution layers.}
      \label{tab:result-ic19} 
\end{table}

\subsection{Ablation Study for Geometric Prior Module}

\subsubsection{The unsupervised optimization objection}
In Geometric Prior Module (GPM), we utilize $L_{landmark}^{align}$ and $L_{landmark}^{div}$ to constrain the training of the Landmark Detection Network. 
As mentioned in Section \ref{section:4.2}, $L_{landmark}^{align}$ is the probability alignment loss to learn equivariance constraint while $L_{landmark}^{div}$ is the diversity loss to prevent the network from learning $K$ identical landmarks. 
To investigate the function of the above loss functions, we perform ablation studies on our ARText dataset, and the experimental results are presented in Table \ref{tab:abalation-loss}. 
Although the main task of training Landmark Detection Network is implemented by $L_{landmark}^{align}$, the presence of $L_{landmark}^{div}$ has a greater impact on model performance. 
Because if only $L_{landmark}^{align}$ exists, the network tends to meet $L_{landmark}^{align}$ constraint by learning $K$ identical landmarks, where the landmarks can not provide effective information for recognition. 
We also conduct experiments on GPM, as is shown in Table \ref{tab:abalation-loss},
when remove GPM, only keep the character feature $F_c$ ((g) in Figure \ref{Geometric Prior Module}) for Graph Reasoning Module, the performance drops about 2.3\%, which shows the effectiveness of landmarks.

\begin{table}[ht]
	\centering
	\vspace{-3pt}

	\renewcommand\arraystretch{1.5}
	\small

	\begin{tabular}{c|c|c|c|c}
		\hline
		\multirow{2}{*}{Methods} & \multicolumn{3}{c|}{Detection} & E2E  \\
		\cline{2-5}
		 & P & R & F &1-NED   \\
		\hline
		DR TS w/ L1 w/o L2 & 91.13& 88.40& 89.51& 73.20	  \\
		\hline
		DR TS w/o L1 w/ L2 &90.72 & \textbf{88.82}& 89.83&73.55	 \\
		\hline
		DR TS w/o GPM &88.93 & 67.37 & 75.48 & 72.38	 \\
		\hline
		DR TS (ours) &\textbf{92.48}&	88.67 &	\textbf{90.53} &	\textbf{74.69}  \\
		\hline
	\end{tabular}

       \caption{Ablation study for unsupervised optimization objection on ARText. The $L_1$ and $L_2$ represent $L_{landmark}^{align}$ and $L_{landmark}^{div}$, respectively. “TS” dentoes TextSpotter. }
       \label{tab:abalation-loss} 
      \vspace{-5pt}
\end{table}

\subsection{Ablation Study for Graph Reasoning Module}
The function of the graph reasoning module is to effectively fuse the character feature and landmark feature. In order to study the effectiveness of this model, we select two other common approaches for comparison: Concatenation and Summation. 
1) For the concatenation case, we directly concatenate the landmark feature (Figure \ref{fig:overall-architecture} (m)) and the character feature (Figure \ref{fig:overall-architecture} (g)),  then pass the fused feature to the Detection Head. 
2) For the summation case, we bilinear interpolation Landmark Map (Figure \ref{fig:overall-architecture} (j)) to the size of Character Feature (Figure \ref{fig:overall-architecture} (g)), and get the mean value of K channels, then sum them to the Character Feature (Figure \ref{fig:overall-architecture} (g)) for subsequent recognition. 

To avoid the performance gains due to the large number of parameters, we make sure that the ablation experiments parameters are not less than DR TextSpotter by adding several layers of convolutions.
The experimental results are shown in Table \ref{tab:ablation-fusion}, and we can draw the following conclusions: 
1). Concatenation is proven to be useful, but not as effective as the graph method in terms of F-measurement and 1-NED. 
2). Although Concatenation performs better on Precision, its parameters far exceed other models, as well as computation volume and memory consumption. 
3). Summation method, however, make performance degrade in terms of 1-NED, even lower than baseline (AE TextSpotter w/o LM). We think the reason may come from that character feature and landmark feature distribute in different feature space, direct summation only leads to an increase in value but does not make any sense in theory. 
4). The GRM achieves the best recognition performance in terms of 1-NED, which verifies the effectiveness of the graph reasoning module.

\begin{table}[ht]
	\centering

	\renewcommand\arraystretch{1.5}
	\small
	\begin{threeparttable}
	\begin{tabular}{c|c|c|c|c|c}
        \hline
		\multirow{2}{*}{Methods} & \multicolumn{3}{c|}{Detection} & E2E &\multirow{2}{*}{Params} \\
		\cline{2-5}
		 & P & R & F &  1-NED \\
		\hline
		Concatenation & \textbf{92.84} &	87.62&	90.15 &74.16&	252.21M \\
		\hline
		Summation & 91.15&	88.41&90.34&71.88	&	236.53M \\
		\hline
		GRM (ours) & 92.48&	\textbf{88.67} &	\textbf{90.53} &	\textbf{74.69}&	236.37M \\
		\hline
	\end{tabular}
      \end{threeparttable}
      \caption{Abalation study for feature fusion method on ARText. 
 }
 	\label{tab:ablation-fusion} 
      \vspace{-5pt}
\end{table}

\subsection{Ablation study on Landmark Numbers} 
To choose the best number of landmarks, we do experiments with $4, 8, 16, 24$ landmarks on our ARText dataset and keep other settings unchanged. As shown in Table \ref{tab:abalation-landmarknum}, the performance improves as the landmark number increases until $K=16$ in terms of 1-NED. 
We think the reason comes from the unsupervised optimization strategy, where the generated landmarks may contain noise. 
As the number of landmarks increases, the noise will accumulate. 
Moreover, if fewer landmarks are utilized, it is not enough to cover all important positions in character. 
Therefore, only the appropriate number of landmarks can make the model work best.

\begin{table}[ht]
	\centering

	\renewcommand\arraystretch{1.5}
	\small
	\begin{tabular}{c|c|c|c|c}
		\hline
		\multirow{2}{*}{Landmark \ Numbers} & \multicolumn{3}{c|}{Detection} & E2E  \\
		\cline{2-5}
		 & P & R & F &1-NED   \\
		\hline
		4 landmarks& 91.04&	88.52 &	90.04 &	73.52  \\
		\hline
		8 landmarks& 92.45 &87.97	&90.21&74.06\\
		\hline
		16 landmarks (ours) &  92.48&	\textbf{88.67} &	\textbf{90.53} &	\textbf{74.69} \\
		\hline
		24 landmarks&\textbf{93.28} & 85.95 & 90.07 & 73.83 \\
		\hline
	\end{tabular}
 	\caption{Ablation study for landmark number on ARText. }
	\label{tab:abalation-landmarknum}  
\end{table}

\subsection{Visualization Results of Landmark Locations}
\begin{figure}[h!]
\centering
\includegraphics[scale=0.5]{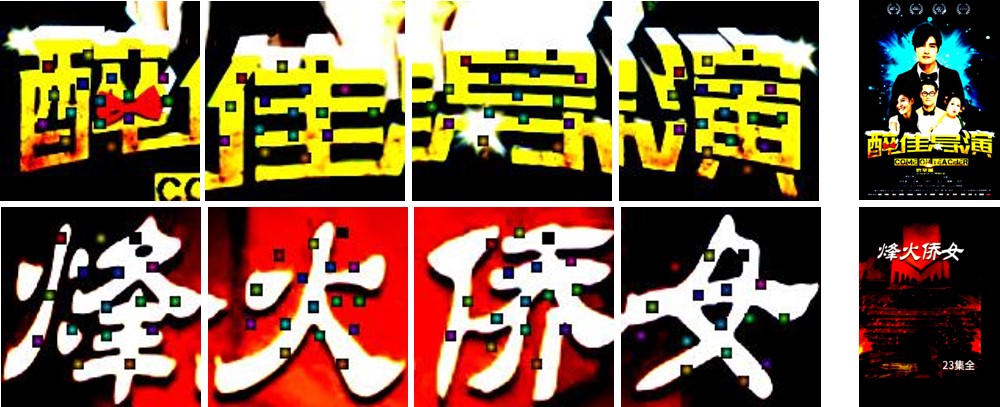}
\vspace{5pt}
\caption{Visualization results of Landmarks. 
Landmarks  can outline the structure of the character, thus providing information to correct wrong recognition due to complex font.}
\label{fig:landmark}
\end{figure}

The visualization results of the landmark are shown in Figure \ref{fig:landmark}. 
The landmarks are mainly distributed in the important positions of the character, such as the start, end and bend positions, thus can provide useful geometric information for the recognition model. 
However, the landmark positions are not particularly accurate, because of the unsupervised learning method.  Actually, how to produce stable and effective landmarks is an open question and our work provides one possible direction. 

\begin{figure*}
\centering
\includegraphics[scale=0.2]{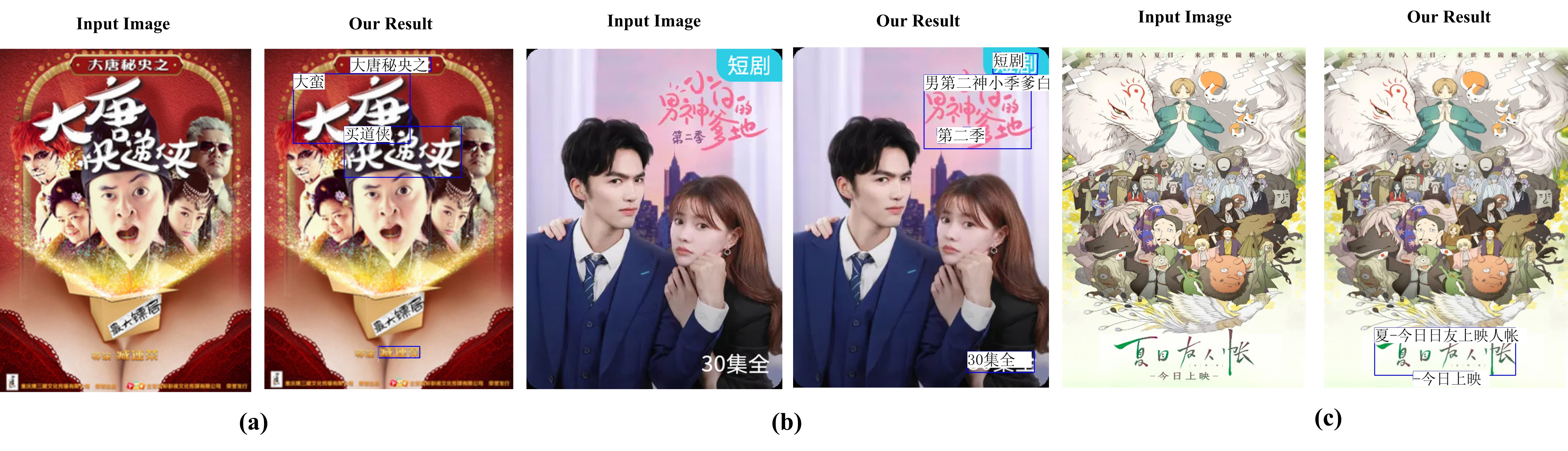}
\captionsetup{singlelinecheck=off,justification=raggedright}
\caption{Failed Examples. (a) is caused by the overlapping between different characters. 
When a part of character crosses with other characters, the model cannot achieve such fine-grained recognition.
(b)(c) are caused by the complex layout. In this case, the model is confused with vertical or horizontal order, resulting in misrecognition.}
\label{fig:failed-example}
\end{figure*}

\begin{figure*}[h!]
\centering
\includegraphics[scale=0.18]{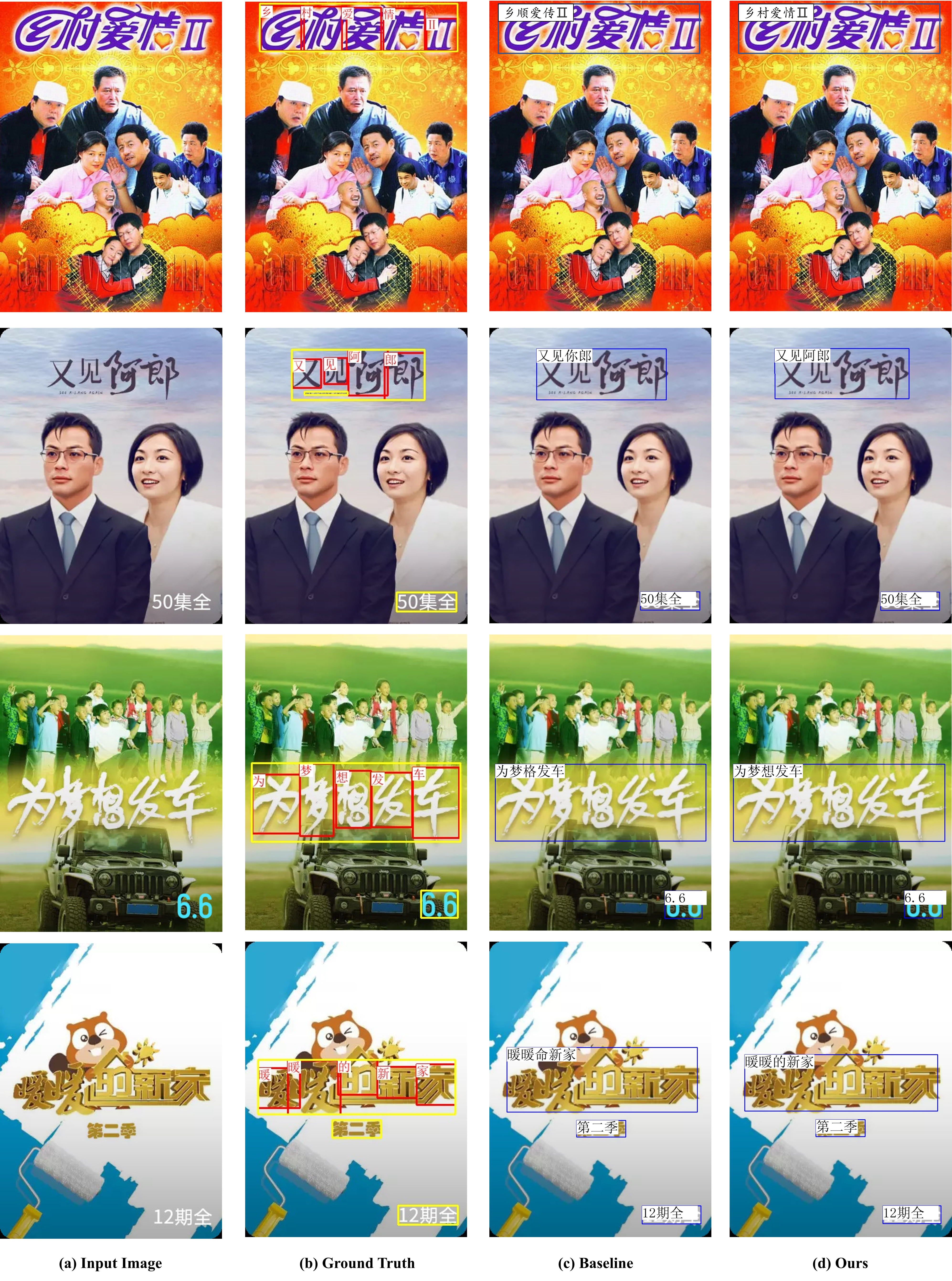}
\caption{Visualization results on ARText dataset. The baseline is AE TextSppoter w/o LM. More examples can be found in Appendix.
}
\label{fig:result}
\end{figure*}

\subsection{Visualization Results on ARText Dataset }
Figure \ref{fig:result} presents the visualization results of the text spotting model. Compared with the baseline model (AE TextSpotter \cite{wang2020ae} without language model), our proposed model can correctly recognize hard characters with complex fonts, which illustrates the robustness of our model to solve the large shape variance problem.

\subsection{Failure Cases and Limitations}
\label{section:failure}
Although our model achieves state-of-the-art performance, it still fails in some special cases.
Some posters are carefully designed in order to highlight the visual effect, thus can result in two main problems: Character overlap and Complex layout, corresponding to Figure \ref{fig:failed-example} (a) and (b)(c).
1. When different text lines are arranged closely or even cross with each other, it will cause difficulties for the model in both detection and recognition. 
2. The complex layout is another serious issue in artistic text spotting. The model cannot arrange each character into the appropriate order, resulting in misrecognition.

Besides, to train our model, we need character-level annotations, which are not easy to transfer to other datasets.
This can be solved by semi-supervised training scheme \cite{xing2019convolutional}: (1) Use the SynthText tool to generate a synthetic dataset with character-level annotations. (2) Pre-train model on the synthetic dataset. (3) Finetune the model on the target dataset with text-level annotations by gradually identifying the “correct” char-level bounding boxes from real-world images by the model itself. Specifically, when finetuning on the real dataset, the model collects the “correct” char-level bounding boxes detected in real-world images, which are used to further train the model.
However, in our model, each character needs to be fed into Character Recognition Branch, including GPM, GRM and Detection Head, which is time-consuming for performing pre-training or semi-supervised schemes to train a text spotting model for the languages (such as English) with many characters in each text instance. 
In summary, for the languages (such as Chinese and Japanese) with few characters in each text instance, even without character-level annotations, we can generate a synthetic dataset with character-level annotations, and then perform semi-supervised scheme to train our text spotting model. 
For the languages (such as English) with many characters in each text instance, training our model under semi-supervised scheme will take a long time, thus the character-level annotations are needed to optimize our model. 

In this paper, we only focus on the font diversity problem in this work, since many plug-and-play modules have been proposed to solve the orientation problem \cite{shi2018aster,2019Efficient,2019Character}, which can be incorporated into our models.

\subsection{Results on English dataset}
\label{section:res-on-engdata}
Based on the discussion in Section \ref{section:failure}, we cannot conduct experiments on common English datasets.
However, to further verify the effectiveness and generalization of our proposed method, 
we select images with English annotations from IC19-ReCTS dataset and set "ignore" label True for Chinese content in the images.
Then we get 10,308 images with 22,135 English text annotations and 147,062 English character annotations. We split into training set and testing set by 8:2.
We train our model and other state-of-the-art methods (without pre-training) on the ReCTS-eng dataset, the results can be seen in Table \ref{tab:result-ic19-eng}.
Our proposed DR TextSpotter outperforms the state-of-the-art method by about 0.9 \% in terms of normalized edit distance, which could verify the effectiveness of our method on the English dataset.

\begin{table}[]
		\centering
	    \renewcommand\arraystretch{2.0}
	    \small

	\begin{threeparttable}
	\begin{tabular}{c|c|c|c|c}
        \hline
	\multirow{2}{*}{Methods} &\multicolumn{3}{c|}{Detection} & E2E  \\
		\cline{2-5}
		 & P & R & F & 1-NED    \\
		\hline
		Mask TextSpotter v3  & 84.12 &73.23&78.30&56.61  \\
		\hline
		AE TextSpotter w/o LM  &84.15&	73.66 &	78.55&	53.84 \\
  		\hline
		SwinTextSpotter &86.29 &\textbf{80.70}&83.40&57.04		\\
		\hline
		DR TextSpotter (Ours) &\textbf{88.96}&	79.36&	\textbf{83.89}&	\textbf{57.96} \\
		\hline
	\end{tabular}
      \end{threeparttable}
      \caption{Results on ReCTS-Eng dataset. “P”, “R”, “F” and “1-NED” mean the precision, recall, F-measure, and the edit-distance-based evaluation metric, respectively. “LM” denotes the language module. E2E means end-to-end text spotting result}
      \label{tab:result-ic19-eng}  
\end{table}

\section{Conclusions}
 The font diversity and shape variance of characters have been ignored in text-spotting tasks for a long time. 
 To remedy this problem, we present a Chinese Artistic Dataset, termed as ARText, which contains $33,000$ artistic images with character-level annotation. 
 Then we propose a deformation robust text spotting method (DR TextSpotter) to detect and recognize text images under the geometric prior. 
 To be concrete, the unsupervised landmark detection sub-network is utilized to highlight the important features as the auxiliary geometric information. 
 Then the graph convolutional network is constructed to fuse this information with the character features, and performs global reasoning to enhance the discrimination of different characters. 
 Experimental results demonstrate the effectiveness of our method on both ARText and IC19-ReCTS datasets. 

\newpage

\bibliographystyle{named}
\bibliography{ijcai23}

\end{document}